# Semi-Supervised Method using Gaussian Random Fields for Boilerplate Removal in Web Browsers

Joy Bose
Microsoft IDC
Hyderabad, India
joy.bose@microsoft.com

Sumanta Mukherjee
Microsoft IDC
Bangalore, India
sumanta.mukherjee@microsoft.com

*Abstract*— Boilerplate removal refers to the problem of removing noisy content from a webpage such as ads and extracting relevant content that can be used by various services. This can be useful in several features in web browsers such as ad blocking, accessibility tools such as read out loud, translation, summarization etc. In order to create a training dataset to train a model for boilerplate detection and removal, labeling or tagging webpage data manually can be tedious and time consuming. Hence, a semi-supervised model, in which some of the webpage elements are labeled manually and labels for others are inferred based on some parameters, can be useful. In this paper we present a solution for extraction of relevant content from a webpage that relies on semi-supervised learning using Gaussian Random Fields. We first represent the webpage as a graph, with text elements as nodes and the edge weights representing similarity between nodes. After this, we label a few nodes in the graph using heuristics and label the remaining nodes by a weighted measure of similarity to the already labeled nodes. We describe the system architecture and a few preliminary results on a dataset of webpages.

*Keywords*— Text classification, Gaussian Random Fields, boilerplate removal, semi supervised learning

## I. Introduction

The problem of removing noisy content such as ads, headers, footers and other clutter from a webpage and extract the relevant content is termed as boilerplate removal [1] or content extraction. It is a problem of text classification, to classify the elements of the webpage as relevant or noise. This is a very important problem for web browsers such as Google Chrome or Microsoft Edge, which have to make the decisions for extracting relevant content in a number of features and services such as reading view [2] and translation service, educational tools such as notes, and accessibility tools such as read out loud.

Various approaches have been proposed to solve this problem of extracting relevant content from web browsers. Such solutions include heuristics, machine learning based approaches for classification such as decision trees, SVM, CRF, and approaches based on webpage layouts, such as VIPS [3-5]. Popular open source and commercial solutions to the problem of boilerplate removal include arc90 readability [6] and boilerpipe [7]. Major web browsers such as Google Chrome and Mozilla Firefox or Edge mainly use heuristics or rule based methods currently, since they are faster, although the accuracy of such methods may not be good for images and embedded elements.

To train a machine learning model to perform this classification task for webpage elements, it is important to create a training and testing dataset of labeled webpages. This involves labeling of each webpage element manually to make the ground truth dataset, which can be a tedious task. This is usually performed by crowdsourcing involving multiple labelers, such as Amazon's Mechanical Turk. However, manual labeling is subjective and can be inconsistent between labelers. Therefore, a semi-supervised approach, where some pages are labeled manually and the system labels the remaining unlabeled elements based on the labeled elements, is more suitable. It is also closer to real world situations where the amount of labeled data available is comparatively low.

In this paper, we present a semi supervised learning method that relies on similarity of webpage elements to label them as noise or non-noise. The approach is inspired by theory of Gaussian random fields, as explained in the work by Zhu and Gaharmani [8].

The rest of the paper is structured as follows: in the following section we preview related work in semi-supervised learning. Section 3 describes the theory behind our approach. Sections 4 outlines the steps for our semi-supervised approach, when applied to classification of webpage elements. Section 5 describes a testing framework for validating our model. Section 6 gives the results of some experiments using the testing framework, and section 7 concludes the paper.

## II. Related work in semi-supervised learning

Semi-supervised learning is arguably more suited to working with real world data than the more common supervised and unsupervised learning. This is because in real world data, some of the data is labeled while most of it is unlabeled. It also has cost implications, since unlabeled data is cheaper and getting humans to label the data is time consuming and often expensive.

Approaches to semi-supervised learning include expectation maximization or EM [12, 13], naïve Bayes [14], conditional random fields [15] as well as Markov random walks [16]. Attempts have also been made to use semi-supervised methods to train deep learning models [17, 18].

Semi-supervised models have been used in a few web related applications such as webpage classification [19] and information retrieval for web search [20]. However prior work is not available in using semi-supervised learning for classifying individual elements within a given webpage. This is the problem this paper focuses on.

In the following sections, we outline the theory of our approach and how it is used in practice for boilerplate removal in webpages.

### III. THEORY

A webpage can be represented as a DOM (Document object model) tree, which shows the relation between different DOM nodes. The DOM nodes in the DOM tree representing the webpage can have text, images or multimedia. For simplification, we only consider leaf nodes of the DOM tree, i.e. those nodes that have no child.

Each such DOM node in a webpage can be represented as an n-dimensional vector $(w_1, w_2, …w_n)$ of various features extracted from the node such as position, dimensions, HTML string, metadata etc. One way of encoding the vectors corresponding to DOM nodes to provide input to the machine learning model, can be using word embeddings, first formulated by Mikolov [4]. Word embeddings use the frequency of neighboring words in the training set to make a vector representation of each word. The word embeddings model is trained to learn the association of the surrounding words to each word as it occurs in the dataset. The Euclidean distance between any two such nodes is can be considered as a representation of similarity.

Let us assume there are n nodes in the webpage $(x_1, y_1), (x_2, y_2),… (x_n, y_n)$ where $x_k$ represents the node k and $y_k$ its label. The labels are binary {0,1} representing 1 for relevant node and 0 for noise node. We assume the first l nodes have labels and nodes (l+1) to n are unlabeled. The labels are provided either by manual effort or by using a set of heuristic rules. The webpage then can be represented in the form of a connected graph G = (V,E) where V is the set of nodes, some of which are labeled and others are unlabeled, and the set of edges E, represented by an adjacency matrix of dimensions n*n, such that the edge $E_{a,b}$ represents the distance between two nodes a and b.

Following the method described by Zhu et al [8, 9], the unlabeled nodes are given continuous valued labels in such a way that the total energy of the graph, defined as the harmonic function (1), is minimized.

$$E = ½ \Sigma w_{ij}(y_i - y_j)^2 \qquad (1)$$

This minimization is done as follows: a Gaussian random field is defined on the graph and a real valued, minimum energy function f is computed on the gaussian random field, and the continuous valued labels on the unlabeled nodes are assigned based on f. The minimum energy function f follows the harmonic property, i.e. its value at each unlabeled node is the average of the values at neighboring nodes.

$$f(j) = 1/d_j \Sigma w_{ij} f(j) \text{ for } j=l+1, l+2, …n \qquad (2)$$

Finally, the continuous labels are made into fixed labels as follows: if the value of the continuous label > 0.5 then it is given a label 1, else it is given 0 as the label.

For each labeled point $x_l$ in the graph, we find the nearest unlabeled point $x_u$, give it the label l, put $x_u$ in the labeled set, and repeat, until all the nodes are labeled.

In the following section, we describe the steps for the semi-supervised learning method for webpage elements. For now, we only work with text nodes in the HTML page, and ignore nodes containing images, multimedia etc.

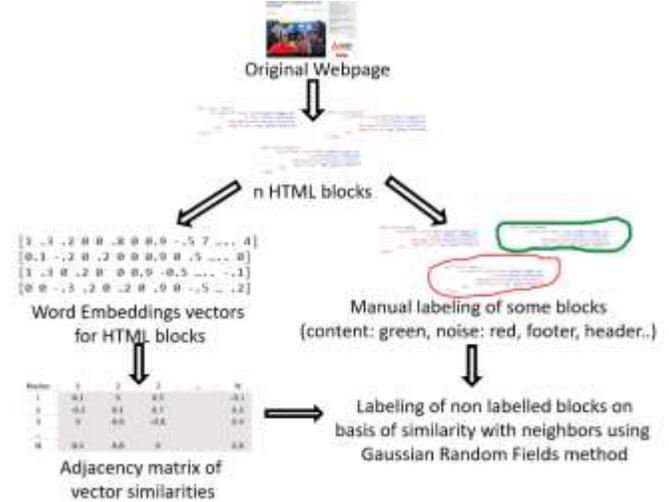

Fig. 1. System for semi supervised labeling of an HTML page based on Gaussian random fields method.

### IV. STEPS FOR SEMI-SUPERVISED LEARNING

Our proposed semi-supervised system for labeling the webpage elements as relevant or not relevant has the following steps:

- Take a webpage as input

- Divide the webpage into HTML blocks of text nodes, by considering only the leaf nodes of the DOM tree and selecting the text nodes. Compute the word embeddings vector of each extracted text node by taking the average of the word embeddings for each word in the text node, generated using a model such as Glove or Word2Vec [10]. Thus, each HTML block is represented as an n-dimensional vector.

- Build the adjacency matrix for graph representation of the webpage, where each HTML block is a node and edge weight is the similarity between nodes.

- Determine the weights of the edges between the nodes in the adjacency matrix, which represents the similarity between nodes and given by the inner product of the vectors for the nodes.

- Generate binary labels for the first l blocks of the webpage into one of the two categories (noise vs

relevant). This labeling can be performed either manually, or using heuristic rules. An example rule is: if a DOM element has <article> tag, then it is relevant content.

- For the unlabeled text nodes, use the previously described method based on harmonic functions and gaussian random fields to label them based on the labels of the neighboring nodes, and applying a threshold. We initialize the function f with 1s or unlabeled nodes and update them in each iteration as described in section 2.

- Continue the previous steps until the whole webpage is labeled.

Fig. 1 illustrates the steps of our method.

## V. TESTING FRAMEWORK

In order to evaluate our semi-supervised algorithm for classification of webpage elements, we built a testing framework. The framework comprises of two components: a tagging tool for labeling webpage elements, and a test bed.

The tagging tool is written in Javascript, as a browser extension. It takes a webpage as input and generates an overlay when the user hovers on a webpage element, which they can then click to select as relevant or not. In this way, the user can manually tag each of the webpage elements. The results are saved in JSON format.

The test bed is based on Selenium, and compares the labeled JSON file output with the output of the extraction algorithm. The accuracy of the algorithm is then evaluated in terms of precision and recall metrics. This step is taken after the manual tagging is complete using the tagging tool and the JSON file is generated. The test bed compares the results of the ground truth dataset with the output of the algorithm described in sections 2 and 3 using Gaussian Random Fields, and outputs the precision and recall values, averaged over the webpages in the dataset.

## VI. EXPERIMENTS AND RESULTS

We built a simple prototype of our system using harmonic functions in Python, using the requests library to get the content for a given URL and BeautifulSoup library to perform parsing of webpage content and extract the text nodes. We used the inbuilt doc vector function of spacy [11] to generate the averaged word embeddings vectors for each sentence.

For the webpage URL, we first extracted the text elements in the HTML using BeautifulSoup, then generated the vectors for each and computed the distance between each of the vectors to construct the adjacency matrix. We manually generated binary labels (noise vs relevant) for 20% of the elements and used the iterative method based to Gaussian functions to generate the remaining labels, whose accuracy we then measured by using human generated labels (tagged previously using our tagging tool) for the same webpage as the ground truth, compared using our test bed.

In our method, we assume that the minimum energy function (described in section 3) is convex and it will converge and generate labels for each text element in the webpage.

We performed the test for 10 URLs, which were news articles from news websites bbc.com, cnn.com and timesofindia.com. We concentrated on news articles for now because they generally have a similar structure and a higher number of readers.

We obtained an accuracy of 70% for the text content extracted from the 10 news article webpages.

Although this result is not perfect, our Gaussian Random fields Semi-Supervised method is still work in progress and its accuracy is expected to improve with better feature selection and other optimizations.

## VII. CONCLUSION AND FUTURE WORK

We have described a semi-supervised method for labeling webpages to remove noisy elements and extract relevant elements. We have built a prototype to evaluate the semi supervised method and obtained some preliminary results. Our method is much more adaptable and requires less labeling effort than the heuristics-based methods used in the browsers currently. In future we intend to improve the method and implement it as a web service.